\documentclass[12pt]{l4dc2020}

\title[Keyframing the Future: Keyframe Discovery for Visual Prediction and Planning]{Keyframing the Future: \\Keyframe Discovery for Visual Prediction and Planning}
\usepackage{times}

\usepackage{amsmath,amsfonts,bm}

\def\eqref#1{equation~\ref{#1}}

\def\1{\bm{1}}

\DeclareMathAlphabet{\mathsfit}{\encodingdefault}{\sfdefault}{m}{sl}
\SetMathAlphabet{\mathsfit}{bold}{\encodingdefault}{\sfdefault}{bx}{n}

\newcommand{\E}{\mathbb{E}}

\usepackage{url}

\usepackage[utf8]{inputenc} %
\usepackage[T1]{fontenc}    %
\usepackage{url}            %
\usepackage{booktabs}       %
\usepackage{amsfonts}       %
\usepackage{nicefrac}       %
\usepackage{microtype}      %

\usepackage{mathtools}
\usepackage{siunitx}
\usepackage{algorithm}
\usepackage{algorithmic}
\usepackage{bm}
\usepackage{wrapfig}
\usepackage{lipsum}
\usepackage{multirow}
\usepackage{xspace}

\definecolor{clr}{cmyk}{0, 0.7808, 0.4429, 0.1412}

\newcommand{\keyin}{\textsc{KeyIn}\xspace}

\makeatletter
\newcommand{\printfnsymbol}[1]{%
  \textsuperscript{\@fnsymbol{#1}}%
}
\makeatother
\newcommand*\samethanks[1][\value{footnote}]{\footnotemark[#1]}

\author{%
 \Name{Karl Pertsch\nametag{\thanks{equal contribution}\thanks{University of Southern California}}} \Email{pertsch@usc.edu}\vspace{-5pt}
 \AND
 \Name{Oleh Rybkin}\nametag{\samethanks[1]\thanks{University of Pennsylvania}}  \Email{oleh@seas.upenn.edu}\vspace{-5pt}
 \AND
 \Name{Jingyun Yang\nametag{\samethanks[2]}} \Email{jingyuny@usc.edu}\vspace{-5pt}
 \AND
 \Name{Shenghao Zhou\nametag{\samethanks[3]}} \Email{shzhou2@seas.upenn.edu}\vspace{-5pt}
 \AND
 \Name{Konstantinos G. Derpanis\nametag{\thanks{Ryerson University and Samsung AI Centre Toronto}}} \Email{kosta@ryerson.ca}\vspace{-5pt}
 \AND
 \Name{Kostas Daniilidis\nametag{\samethanks[3]}} \Email{kostas@seas.upenn.edu}\vspace{-5pt}
 \AND
 \Name{Joseph J. Lim\nametag{\samethanks[2]}} \Email{limjj@usc.edu}\vspace{-5pt}
 \AND
 \Name{Andrew Jaegle\nametag{\thanks{DeepMind, work conducted while still at UPenn}}} \Email{ajaegle@upenn.edu}
}

\begin{document}

\maketitle

\begin{abstract}

Temporal observations such as videos contain essential information about the dynamics of the underlying scene, but they are often interleaved with inessential, predictable details. One way of dealing with this problem is by focusing on the most informative moments in a sequence. We propose a model that learns to discover these important events and the times when they occur and uses them to represent the full sequence.  We do so using a hierarchical Keyframe-Inpainter (\textsc{KeyIn}) model that first generates a video's keyframes and then inpaints the rest %
by generating the frames at the intervening times. We propose a fully differentiable formulation to efficiently learn this procedure. We show that \keyin finds informative keyframes in several datasets with different dynamics and visual properties. \keyin outperforms other recent hierarchical predictive models for planning. For more details, please see the project website at \url{https://sites.google.com/view/keyin.}

\end{abstract}

\section{Introduction}

When thinking about
the future, humans focus %
on the important things that may happen (When will the plane depart?%
) without fretting about the minor details that fill each intervening moment (%
What is the last word I will say to the taxi driver?).
Because the vast majority of elements in a temporal sequence contain redundant information, a temporal abstraction can make reasoning and planning both easier and more efficient. How can we build such an abstraction? Consider the example of a lead animator who wants to draw the next scene of a cartoon. Before worrying about every low-level detail, the animator first sketches out the story by \textit{keyframing}, drawing the moments in time when the important events occur. The scene can then be easily finished by other animators who fill in the rest from the story laid out by the keyframes. In this paper, we argue that learning to discover such informative keyframes%
is an efficient and powerful way to learn to reason about the future.

\begin{figure*}[t]
  \centering
  \includegraphics[width=\linewidth]{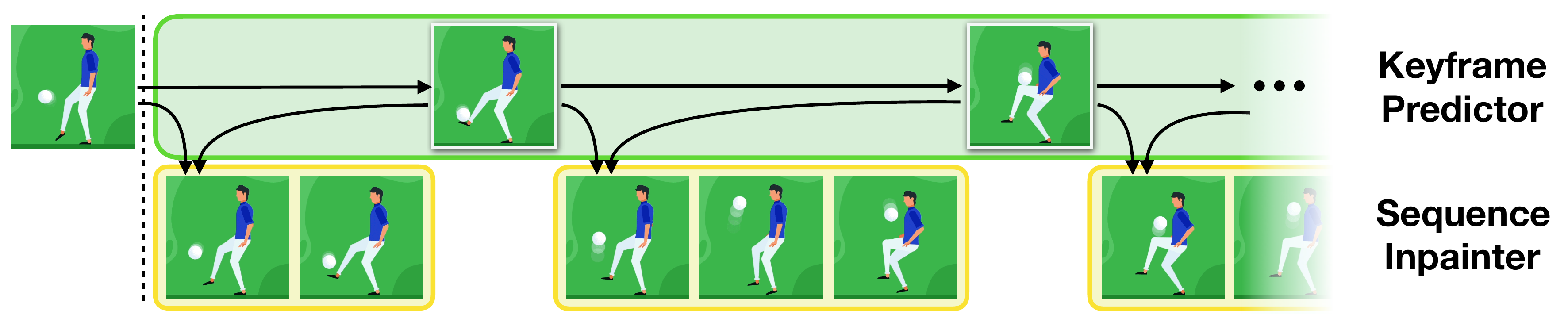}
  \vspace{-20pt}
  \caption{%
  Keyframing the future.
  Instead of predicting one frame after the other, we propose to represent the sequence with the \textit{keyframes} that depict the interesting moments of the sequence. The remaining frames can be inpainted given the keyframes.
  }
  \label{fig:teaser}
  \vspace{-20pt}
\end{figure*}

Our goal is to learn such an abstraction for learning dynamics of images. In contrast, much of the work on future image prediction and planning has focused on frame-by-frame synthesis (\cite{Oh:2015:AVP:2969442.2969560, finn2016unsupervised, ebert2018visual}).
This strategy puts an equal emphasis on each frame, irrespective of
the redundant content it contains or how useful it is for reasoning relative to other predicted frames.
Other recent work has considered predictions that ``jump'' more than one step into the future, using either fixed-offset jumps \citep{buesing2018learning} or heuristics such as predictability of the frame \citep{neitz2018adaptive, Jayaraman2018, gregor2018temporal} to choose which frames to predict. In this work, we instead propose a method that predicts the keyframes that are most informative about the full sequence. We do so by ensuring that the full sequence can be recovered from the keyframes with an \textit{inpainting} strategy, similar to how a supporting animator fleshes out the story keyframed by the lead (see Fig.~\ref{fig:teaser}). The keyframe structure allows us to reason about the sequence holistically when planning future actions while only using a small subset of the frames. Visual model-predictive control (MPC) methods that reason about every single future time step scale poorly if the task requires long-horizon planning. In contrast, our method enables visual planning over much greater horizons by using keyframes as subgoals in a hierarchical planning framework.

Our contributions are as follows. We formulate a hierarchical approach for the discovery of informative keyframes using joint keyframing and inpainting (\textsc{KeyIn}), and propose a soft objective that allows us to train the model in a fully differentiable way. We also propose a hierarchical planning algorithm for this model. We first analyze our model in a controlled setting %
to show that it can reliably recover the underlying keyframe structure on visual data. We then show that our model discovers hierarchical temporal structure on more complex datasets of demonstrations: an egocentric gridworld environment and a simulated robotic pushing dataset, which is challenging for current approaches to visual planning. %
Our approach outperforms existing hierarchical and non-hierarchical planning schemes on the pushing task, enabling long-horizon, hierarchical control.

\section{Related work}

\paragraph{Video modeling.}

Early deep probabilistic video models include autoregressive models that predict the pixels sequentially \citep{kalchbrenner2017video, reed2017parallel}. %
To reason about the images holistically, latent variable approaches were developed based on variational inference \citep{chung2015recurrent, rezende2014stochastic, kingma2014auto}, including \citep{babaeizadeh2018stochastic,denton2018stochastic,lee2018savp,li2018disentangled,he2018probabilistic} and large-scale models such as \citep{castrejon2019improved, villegas2019high}. %
We show how latent variable models can be used to learn temporal abstractions with a novel keyframe-based generative model. %

\paragraph{Visual planning and model predictive control.}
Several groups \citep{Oh:2015:AVP:2969442.2969560, finn2016unsupervised, ChiappaRWM17} have proposed models that predict the future image observations given the agent's actions. \citet{byravan2017se3, hafner2018learning, ebert2018visual} have shown that visual model predictive control based on such models can be applied to a variety of different settings. \cite{fang2019dynamics} shows that a simple jumpy hierarchical prediction method improves planning performance in the real world.  Concurrently, \cite{nair2019hierarchical} design a hierarchical planning method that finds subgoals via extensive planning. In this work, we show that the hierarchical representation of a sequence in terms of keyframes allows more efficient hierarchical planning.

\paragraph{Hierarchical temporal structure.}

Recently, several neural methods were proposed to leverage temporal structure in video data for prediction. \cite{neitz2018adaptive} and \cite{Jayaraman2018} proposed models that find and predict the least uncertain ``bottleneck'' frames. %
\cite{kipf2018compositional} propose a related method for video segmentation via generative modeling, and %
use it for %
hierarchical reinforcement learning. %
\cite{kim2019variational} propose a method for learning temporal abstractions through %
hierarchical state-space models. Concurrently to our work, \citet{shang2019learning} propose a keyframing method that learns to select frames that are informative about the action trajectory.
In contrast to these works, \keyin discovers informative keyframes via joint keyframing and inpainting.

\section{Keyframing the future}

\begin{wrapfigure}{R}{0.55\textwidth}
  \centering
  \vspace{-25pt}
  \includegraphics[width=1\linewidth]{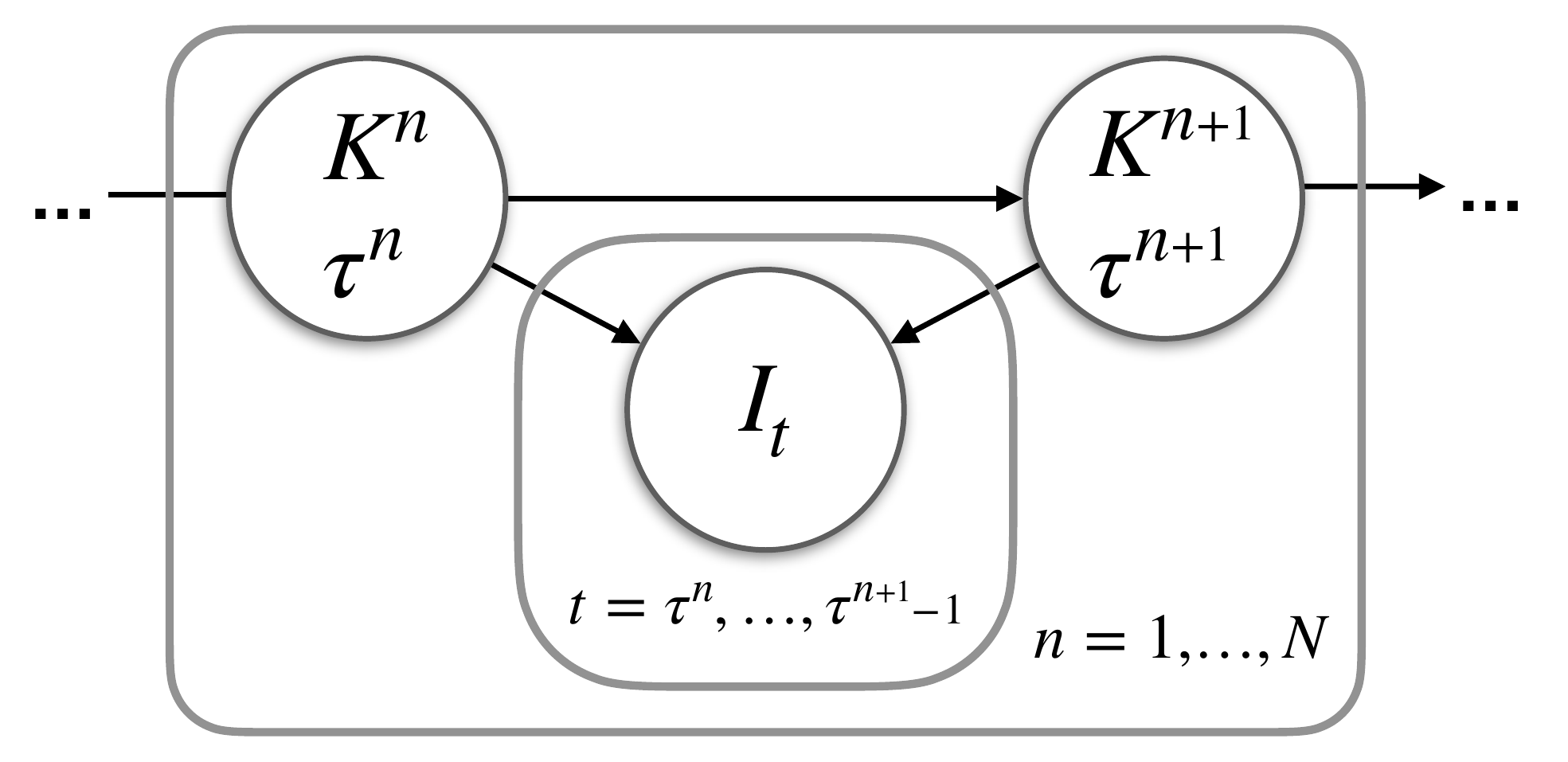}
  \vspace{-15pt}
  \caption{
  A probabilistic model for jointly keyframing and inpainting a future sequence. First, the model generates a sequence of keyframes ${K}^{1:N}$ and the corresponding temporal indices $\tau^{1:N}$ defining the structure of the underlying sequence. Second, the model inpaints the frames $I_{\tau^n:\tau^{n+1}-1}$ for each pair $K^n$ and $K^{n+1}$. %
  }
  \label{fig:pgm}
  \vspace{-20pt}
\end{wrapfigure}

Our goal is to develop a model that generates sequences by first predicting important observations (keyframes) and the time steps when they occur and then filling in the observations in between. To achieve this goal, in the following we (i) define a probabilistic model for joint keyframing and inpainting, and (ii) show how a maximum likelihood objective for this model leads to the discovery of keyframe structure.

\subsection{A probabilistic model for joint keyframing and inpainting}
\label{sec:pgm}

To represent a sequence $I_{1:T}$ via a small set of keyframes, we propose a probabilistic model of the sequence that consists of two parts: %
the \emph{keyframe predictor} %
and the \emph{sequence inpainter} (see Fig.~\ref{fig:pgm}). %

The \emph{keyframe predictor} takes in  $C$ conditioning frames $I_{co}$ and produces $N$ keyframes $K^{1:N}$ as well as the corresponding time indices  $\tau^{1:N}$. It factorizes in time as:
\begin{align}\begin{split}
p(K^{1:N}, \tau^{1:N} |  I_{co}) = \prod_n p(K^n,  \tau^{n} | K^{1:n-1},  \tau^{1:n-1}, I_{co}).
\end{split}\end{align}
From each pair of keyframes, the \emph{sequence inpainter} generates the sequence of  frames in between:
\begin{align}\begin{split}
p(I_{\tau^n:\tau^{n+1}-1} |  K^{n}, K^{n+1}, \tau^{n+1} - \tau^n) = \prod_t p(I_t | K^{n}, K^{n+1}, I_{\tau^n:t-1}, \tau^{n+1} - \tau^n),
\end{split}\end{align}
which completes the generation of the full sequence. The inpainter additionally observes the number of frames it needs to generate $\tau^{n+1} - \tau^n$. The temporal spacing of the most informative keyframes is data-dependent: shorter keyframe intervals might be required in cases of rapidly fluctuating motion, while longer intervals can be sufficient for steadier motion. Our model handles this by predicting the keyframe indices $\tau$ and inpainting $\tau^{n+1} - \tau^n$ frames between each pair of keyframes. %

\subsection{Keyframe discovery}

If a simple model is used for inpainting, most of the representational power of the model has to come from the keyframe predictor. We use a powerful stochastic latent variable model for the keyframe predictor and a simpler predictor network without stochastic latent variables for inpainting. Because of this structure, the keyframe predictor has to predict keyframes that describe the underlying sequence well enough to allow a simpler inpainting process to maximize the likelihood. %

Specifically, to produce a complex multimodal distribution over $K$ we use a per-keyframe latent variable $z$ with prior distribution $p(z)$ and approximate posterior $q(z | I, I_{co})$.\footnote{For simplicity, the variable representing the full sequence is written without indices ($I$ is the same as $I^{1:T}$).} We construct a variational lower bound on the likelihood of both $I$ and $K$ as follows:%
\begin{align}\begin{split}
   \ln  p(I, K |  I_{co})  \geq & \;
  \mathbb{E}_{q(z | I, I_{co})} \bigg[
   \sum_{n=1}^N \underbrace{\ln \E_{p(\tau^{n}, \tau^{n+1}|z^{1:n},  I_{co})} \left[ p(I_{\tau^n:\tau^{n+1}} |  K^{n, n+1}, \tau^{n+1} - \tau^n) \right]}_{\text{inpainting}}
  \\
  & +
  \underbrace{\ln p(K | z,  I_{co})}_\text{keyframing} \bigg]
  - \underbrace{D_{\text{KL}}\left(q(z | I, I_{co})||p(z)\right)}_\text{regularization}.
 \label{eq:elbo}
\end{split}\end{align}

In practice, we use a weight $\beta$ on the KL-divergence term, as is common in amortized variational inference~\citep{higgins2017beta, pmlr-v80-alemi18a, denton2018stochastic}.

\section{Continuous relaxation by linear interpolation in time}
\label{sec:cont_relax}

In principle, this model can dynamically predict the keyframe placement $\tau^n$.
However, learning a distribution over the discrete variable $\tau^n$ is challenging due to the expensive evaluation of the expectation over $p(\tau^{n}|z^{1:n},  I_{co})$ in Eq. \ref{eq:elbo}. To learn the keyframe placement efficiently and in a differentiable manner, we propose a continuous relaxation of the objective. %

\paragraph{Keyframe targets.} Instead of sampling from $\tau^n$ to pick a target frame, we compute the \emph{expected} target frame %
$\tilde{K}^{n}$ - we linearly interpolate between the ground truth images according to the predicted distribution over the keyframe's temporal placement $\tau^n$: $\tilde{K}^{n} = \sum_t \tau^n_t I_{t}$, where $\tau^n_t$ is the probability that the $n^{th}$ keyframe occurs at timestep $t$ (see supplement, Fig.~\ref{fig:loss}).
When the entropy of $\tau^n$ converges to zero, the resulting continuous relaxation objective is equivalent to the original, discrete objective. \footnote{We find this occurs most of the time in practice.}

We parametrize temporal placement prediction in terms of offsets $\delta$ with a maximum offset of $J$. The maximum possible length of the predicted sequence is then $NJ$. Large values of $J$ may allow the model more flexibility, but this may also %
lead to the generation of sequences longer than the target $NJ>T$. %
To force the model to predict valid sequences at training time , we discard predicted frames at times $> T$ and normalize the placement probability over the first $T$ steps. Specifically, for each keyframe we compute this probability as $c^n$: $c^n = \sum_{t \leq T} \tau^n_{t}$.
The loss corresponding to the last two terms of Eq.~(\ref{eq:elbo}) then becomes:
\begin{equation}
    \mathcal{L}_{key} = \frac{\sum_{n} c^n \left(%
    || \hat{K}^n -  \tilde{K}^{n} ||^2
    + %
    \beta\;D_{\text{KL}}\left(q(z^{n} | I, I_{co}, z^{1:n-1})||p(z^{n})\right) \right)}{ \sum_{n} c^n}.
\end{equation}

\paragraph{Inpainting targets.} %
We produce a target image composed from the inpainted frames for each \textit{ground truth} frame.\footnote{This ensures that each ground truth frame contributes equally to the final loss.
}
We note that as offsets $\delta$ have a maximum range of $J$, and in general have non-zero probability on each timestep, the inpainting network needs to produce $J$ frames $\hat{I}^n_{1:J}$  between each pair of keyframes $\big(K^n, K^{n+1}\big)$.
The expected targets are computed as: %
$I_t$: $\tilde{I}_t = (\sum_{n,j} m^n_{j,t} \hat{I}^{n}_j) / \sum_{n,j} m^n_{j,t}$. Here, $m^n_{j,t}$ is the probability that the $j$-th predicted image in segment $n$ has an offset of $t$ from the beginning of the predicted sequence, which can be computed from $\tau^n$. To obtain a probability distribution over produced frames, we normalize the result with $\sum_{n,j} m^n_{j,t}$.

A detailed description of the loss computation can be found in the supplement, Sec.~\ref{sec_supp_loss}. The full loss for our model is:
\begin{equation}
\label{eq:loss}
\mathcal{L}_{total} = \mathcal{L}_{key} + \beta_{I}\sum_{t} || I_t - \tilde{I}_t ||^2.
\end{equation}

\section{Keyframe-based planning}
\label{sec:planning_procedure}

\begin{wrapfigure}{R}{0.50\textwidth}
  \centering
  \includegraphics[width=1\linewidth]{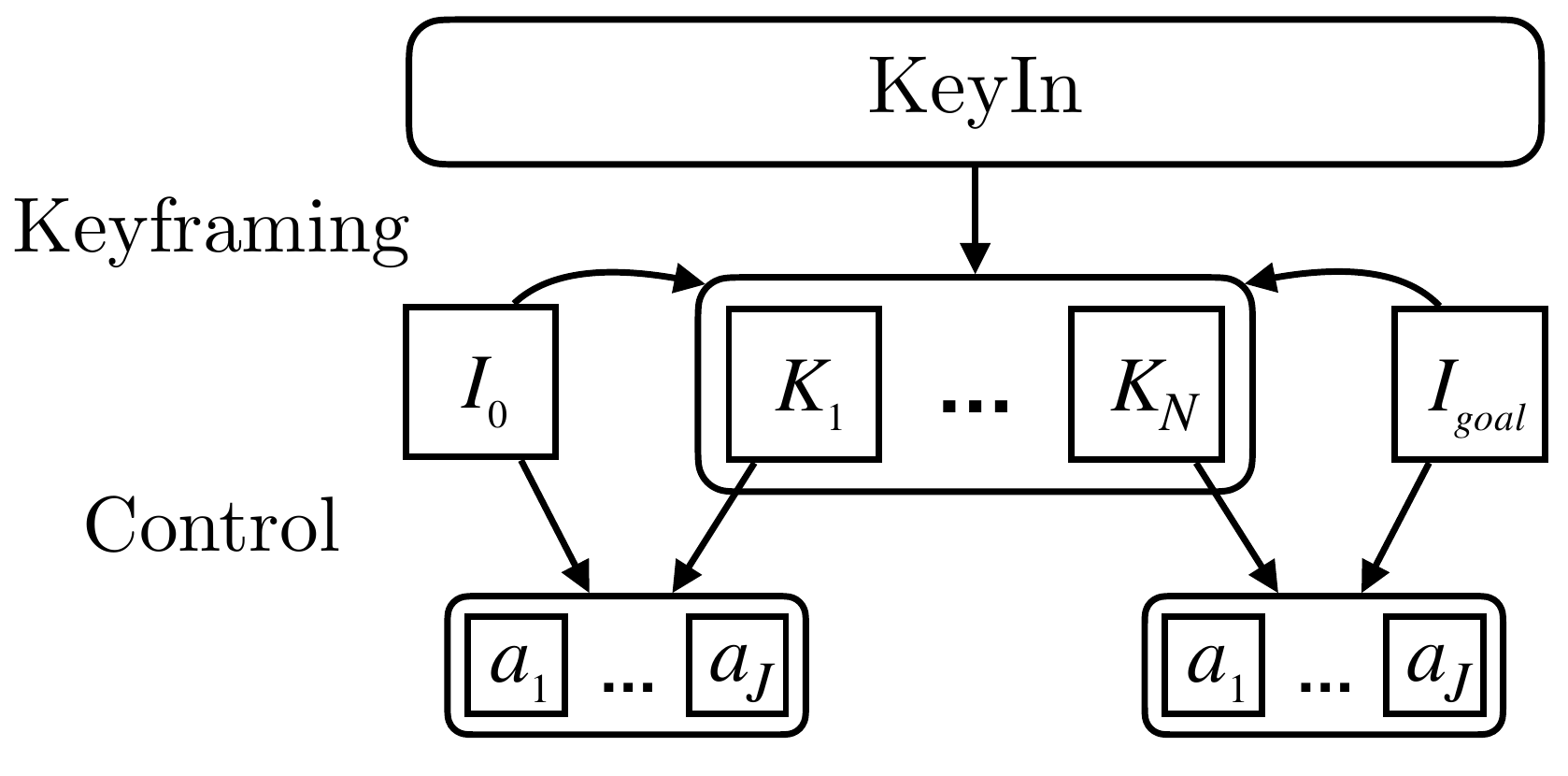}
  \vspace{-20pt}
  \caption{
  Keyframe-based planning. We use the keyframe model to plan a sequence of keyframes between the current observation image and the goal. A low-level controller, e.g. based on model predictive control, produces the actions, $a_t$, executed to reach each keyframe, until the final goal is reached.
  }
  \label{fig:plan_method}
  \vspace{-20pt}
\end{wrapfigure}

We next describe how we use the keyframe-based prediction model for long-horizon, keyframe-based planning. The hierarchical planning procedure is outlined in Fig.~\ref{fig:plan_method}. %
We can generate keyframe trajectories $\hat{K}^{1:N}$ from our model by rolling out trajectories of latent variables $z$ sampled from a Gaussian distribution $\mathcal{N}({\mu}, {\sigma})$. The planning problem can be formalized as finding the set of latent variables $z^{\ast}$ for which the resulting keyframe trajectory minimizes a given cost function $c$, e.g. the final distance to the goal image: $\min_z c(\hat{K}^{N}(z), I_{\text{goal}})$. To optimize this objective, we use the Cross-Entropy Method (CEM, \cite{blossom2006cross}), which is conceptually simple and has given good results in similar settings in prior work (\cite{hafner2018learning, ebert2018visual}). CEM is a sampling-based optimizer that iteratively refits the sampling distributions to those parts of the latent space that resulted in trajectories of low cost. We describe the CEM procedure in more detail in the supplement, Sec.~\ref{sec_supp_planning} and provide details on the used cost function in the experimental section.

After planning a sequence of subgoals towards the goal, we execute the plan by sequentially reaching the subgoals using a low-level controller. \keyin is agnostic to the choice of low-level controller used to reach the intermediate goals. %

\section{Experiments}

\begin{figure*}
  \centering
  \vspace{-4pt}
  \includegraphics[width=\textwidth]{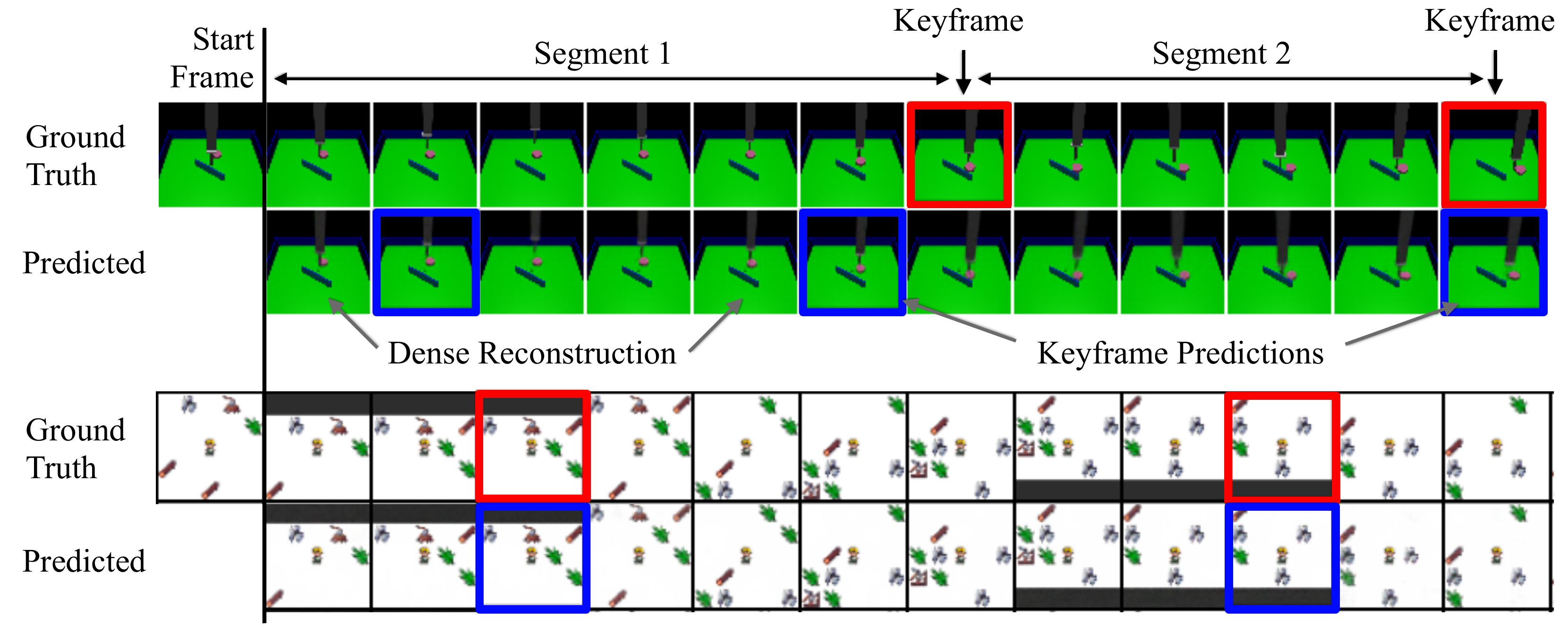}
  \vspace{-25pt}
  \caption{Example generations by \textsc{KeyIn}. %
  The generation is conditioned on a single ground truth frame. Twelve of the 30 predicted frames are shown. %
  We observe that for each transition between pushes and each action of the Gridworld agent our network predicts a keyframe either exactly at the timestep of the event or one timestep apart. %
  }
  \label{fig:push_seqs}
   \vspace{-25pt}
\end{figure*}

We evaluate the quality of \keyin's representation for future sequences by addressing the following questions: (i)~Can it discover and predict informative keyframes? (ii)~Can it model complex data distributions? (iii)~Is the discovered hierarchy useful for long-horizon hierarchical planning?

We instantiate KeyIn using neural networks and train our model using a two-stage training procedure in which we first train the sequence inpainter to inpaint between ground truth frames sampled with random offsets and then train the keyframe predictor with the loss from Eq.~\ref{eq:loss} while freezing the weights of the inpainter. We found this lead to improved results. For further details on model architecture, training and hyperparameters we refer to the supplement, Sec.~\ref{supp_architecture_training}~\&~\ref{sec_supp_experimental} and Fig~\ref{fig:architecture_all}.

\paragraph{Datasets.} We evaluate our model on three datasets containing structured long-term behavior. The \emph{Structured Brownian motion}~(SBM) dataset consists of binary image sequences of size $32 \times 32$ pixels in which a ball randomly changes directions after periods of straight movement of six to eight frames. %

The \emph{Gridworld Dataset} consists of 20k sequences of an agent traversing a maze %
with different objects. The agent sequentially navigates to objects and interacts with them. %
We use the same maze for all episodes and randomize the initial position of the agent and the task sketch. %
We use $64 \times 64$~pixel image observations and further increase the problem complexity by constraining the field of view to a $5\times 5$~cells egocentric window. %

The \emph{Pushing Dataset} consists of 50k sequences of a robot arm pushing a puck towards a goal on the opposite side of a barrier. Each sequence consists of six consecutive pushes. %
We vary start and target position of the puck, as well as the placement of the barrier. %
The demonstrations were generated with the MuJoCo simulator \citep{todorov2012mujoco} at a resolution of $64 \times 64$~pixels. For more details on the data generation process, see supplement, Sec.~\ref{sec_supp_mjc_data}.

\subsection{Keyframe discovery}
\label{sec:keyframe_discovery}

\begin{wraptable}{R}{0.6\textwidth}
\vspace{-25pt}
\caption{F1 accuracy score for keyframe discovery on all three datasets. Higher is better.}\label{tab:F1_kf_disc}
\vspace{-20pt}
\begin{center}
\begin{small}
\begin{sc}
\begin{tabular}{llll}
\toprule
Method                     & Brownian & Push & Gridworld \\
\midrule
Random                     & $0.15$ %
                           & $0.18$ %
                           & $0.12$
                \\
Static                     & $0.21$                   & $0.18 $       & $0.25$\\
Surprise                    & $0.73$                   & $0.17 $     & $0.32$  \\
KeyIn (Ours) & $\bm{0.94}$        & $\bm{0.43}$    & $\bm{0.42}$\\
\bottomrule
\end{tabular}

\end{sc}
\end{small}
\end{center}
\vspace{-25pt}
\end{wraptable}

To evaluate \textsc{KeyIn}'s ability
to discover keyframes, we train \keyin  on all three datasets with $N=6$ keyframes, which can be interpreted as selecting the $N$ most informative frames from a sequence.  We show qualitative examples of keyframe discovery for the Gridworld and Pushing datasets in Fig.~\ref{fig:push_seqs} and for the SBM dataset in the supplement, Fig.~\ref{fig:bb_baseline}. %

\begin{wrapfigure}{R}{0.5\textwidth}
\vspace{-10pt}
\includegraphics[width=\linewidth]{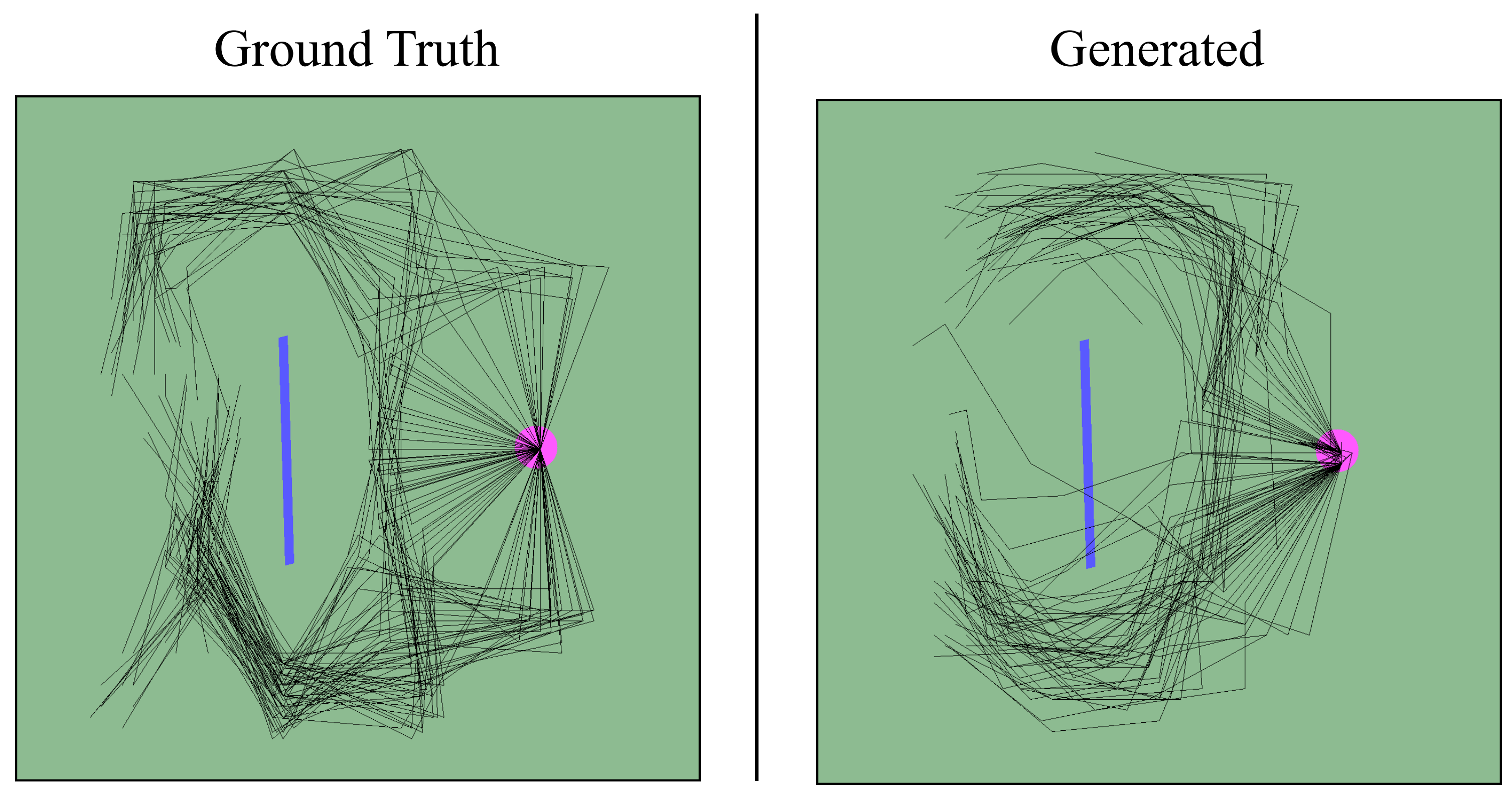}
 \vspace{-20pt}
\caption{Distribution of trajectories sampled from \keyin. Each black line denotes one of 100 trajectories of the manipulated object. The barrier is shown in blue, initial position in pink. %
The model covers both modes of the distribution. %
}
   \label{fig:push_kf_disc}
  \vspace{-15pt}
\end{wrapfigure}

For quantitative analysis, we define approximate ground truth keyframes to be the points of direction change for the SBM dataset, the moments when the robot lifts its arm to transition between pushes, or when the agent interacts with objects in the gridworld. We report F1 scores that capture both the precision and recall of keyframe discovery. We compare to random keyframe placement, a learned but static baseline that chooses identical keyframe placement for all sequences, and a method based on surprise that is similar to prior approaches (see supplement, Sec.~\ref{sec_surprise}). The evaluation in Table~\ref{tab:F1_kf_disc} shows that \keyin discovers better keyframes than alternative methods. %
We analyze the robustness of keyframe discovery to misspecified number of keyframes and image noise in supplement, Sec.~\ref{sec:supp_kf_robust}.%

\subsection{Keyframe-based video modeling}

We verify that \textsc{KeyIn} can represent complex data distributions in terms of discovered keyframes and attains diversity and visual quality comparable to state-of-the-art prediction models. We show sample generations on the Pushing and Gridworld datasets on the supplementary website. %
Fig.~\ref{fig:push_kf_disc} visualizes multiple sampled Pushing sequences from our model conditioned on the same start position, showing that \keyin is able to cover both modes of the demonstration distribution. We further show that \keyin is competitive with prior approaches on video prediction metrics for sequence modeling and outperforms prior approaches in terms of keyframe modeling in Tables~\ref{tab:ssim_psnr}~\&~\ref{tab:kf_ssim_psnr} in the supplement.

\subsection{Hierarchical keyframe-based planning} \label{sec:planning}

We test whether the inferred keyframes can be used as subgoals for hierarchical planning in the pushing environment. %
We follow the planning procedure detailed in Sec.~\ref{sec:planning_procedure}. We design a simple cost function for the pushing domain based on detected centroids of the puck in both the goal image and the predicted keyframes (more details in supplement, Sec.~\ref{sec_supp_planning}). After finding a plan of subgoals, a low-level controller reaches each subgoal via model predictive control using ground truth dynamics, employing CEM for optimization of the action trajectory.

We find that \keyin is able to plan coherent subgoal paths towards the final goal that often lead to successful task execution (executions are shown on the supplementary website\footnote{\url{https://sites.google.com/view/keyin}} and in the supplement, Fig.~\ref{fig:push_exec_traces}).
To quantitatively evaluate the keyframes discovered, we compare to alternative subgoal selection schemes: fixed time offset (\emph{Jumpy}, similar to \citet{buesing2018learning}), a method that determines points of peak surprise (\emph{Surprise}, see Sec.~\ref{sec:keyframe_discovery}), a bottleneck-based subgoal predictor (time-agnostic prediction or TAP, \citet{Jayaraman2018}), and subgoals selected at fixed intervals from sequences generated by CIGAN, an alternative sequence modeling approach (\cite{wang2019learning}). We additionally compare to an approach that plans directly towards the final goal using the low-level controller (\emph{Flat}).
We evaluate all methods with the shortest path between the target and the actual position of the object after the plan is executed.
As the goal of this experiment is to evaluate the quality of predicted subgoals, all methods use the same low-level controller.

\begin{wraptable}{R}{0.53\textwidth}
\vspace{-20pt}
    \caption{Planning performance on the Pushing task. }
        \label{tab:planning}
        \vspace{-20pt}
        \begin{center}
        \begin{small}
        \begin{sc}
        \resizebox{0.5\textwidth}{!}{
        \begin{tabular}{lcc}
        \toprule
        Method           & Position error & Success Rate \\
        \midrule
        Intitial         & $1.32 \pm 0.06$       & -    \\
        Random           & $1.32 \pm 0.07$       & -   \\
        \midrule
        Flat      & $0.90 \pm 0.14$       & \SI{15.0}{\percent}   \\
        TAP              & $0.80 \pm 0.16$       & \SI{23.3}{\percent}    \\
        Surprise          & $0.64 \pm 0.28$       & \SI{50.8}{\percent}    \\
        Jumpy            & $0.62 \pm 0.33$       & \SI{58.8}{\percent}    \\
        Jumpy - CIGAN    & $0.99 \pm 0.19$       & \SI{15.8}{\percent}   \\
        KeyIn (Ours)     & $\bm{0.50 \pm 0.26}$  & {\textbf{\SI{64.2}{\percent}}}     \\
        \bottomrule
        \end{tabular}
        }
        \end{sc}
        \end{small}
        \end{center}
\vspace{-25pt}
\end{wraptable}

As shown in Table~\ref{tab:planning}, %
our method outperforms prior approaches. %
TAP shows only a moderate increase in performance over the Flat planner, which is likely because it fails to predict good subgoals and often simply predicts the final image as the bottleneck. We think this is due to the large stochasticity of our dataset and the absence of the clear bottlenecks that TAP is designed to find. Our method outperforms the planners that use Jumpy and Surprise subgoals. This further confirms that \keyin is able to produce informative keyframes, such that it is easier for the low-level controller to follow them. %
\vspace{-7pt}

\section{Conclusion}
\vspace{-3pt}

We introduced \keyin, a method for representing a sequence through its informative keyframes by jointly keyframing and inpainting. \keyin first generates the keyframes of a sequence and their temporal placement and then produces the full sequence by inpainting the frames in between. We showed that \keyin discovers informative keyframes on several datasets with stochastic dynamics. Furthermore, by using the keyframes for planning, we showed our method outperforms several other hierarchical planning schemes. %

\subsubsection*{Acknowledgements}
We thank the members of the GRASP laboratory at Penn, CLVR laboratory at USC, and RAIL at UC Berkeley for many fruitful discussions. We thank Marvin Zhang, Leonard Hasenclever and Shao-Hua Sun for helpful comments on the manuscript, Dinesh Jayaraman for help with the TAP baseline, and Frederik Ebert for advice on CEM planning. We are grateful for support through the following grants: NSF-IIP-1439681 (I/UCRC), NSF-IIS-1703319, NSF MRI 1626008, ARL RCTA W911NF-10-2-0016, ONR N00014-17-1-2093, ARL DCIST CRA W911NF-17-2-0181, the DARPA-SRC C-BRIC, and by Honda Research Institute. K.G.D. is supported by a Canadian NSERC Discovery grant. K.G.D. contributed to this work in his personal capacity as an Associate Professor at Ryerson University.

\clearpage
\clearpage

\bibliography{bibref_definitions_long,bibtex}

\clearpage
\appendix

\section{Deep Video Keyframing}
\label{supp_architecture_training}

\begin{figure}
  \centering
  \includegraphics[width=\linewidth]{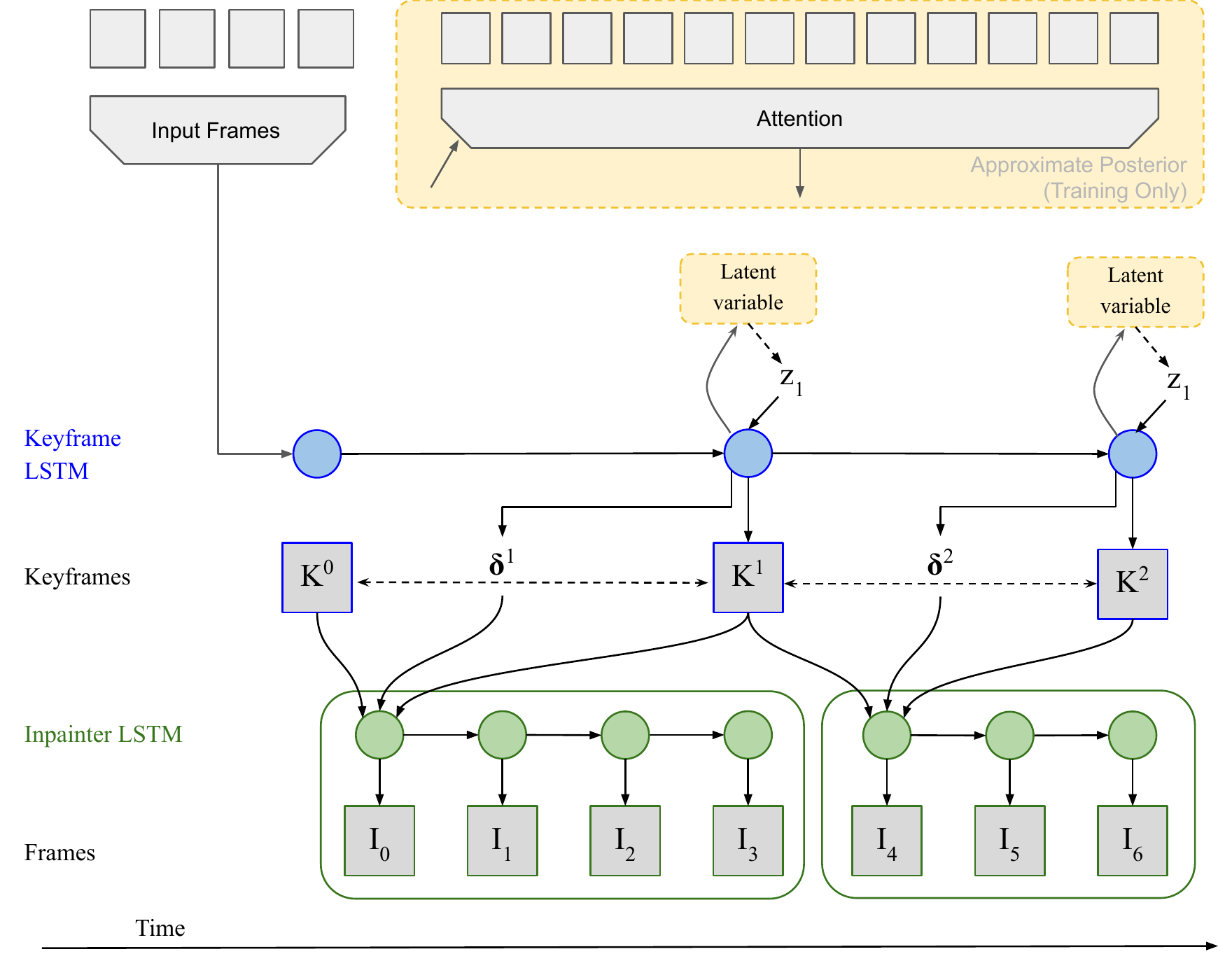}
  \vspace{-20pt}
  \caption{Overview of the model architecture. The Keyframe predictor is an LSTM network that predicts keyframes as well as the offsets between keyframes. It is conditioned on a sequence of input frames. For each pair of frames and the offset between them, the inpainter network produces the intermediate predictions. The keyframe predictor is formulated as a latent variable model. We optimize the entire model with variational inference, where at training the latents $z$ are produced from a variational posterior, and at test time simply sampled from the prior. The variational posterior is implemented as an LSTM with attention over the future frames. 
  }
  \label{fig:architecture_all}
  \vspace{-15pt}
\end{figure}

We show how to instantiate \keyin with deep neural networks and train it on high-dimensional observations, such as images. We further describe an effective training procedure for \keyin. The overview of the model is shown in Figure \ref{fig:architecture_all}.

\subsection{Architecture}%
\label{sec_supp_architecture}

We use a common encoder-recurrent-decoder architecture (\cite{denton2018stochastic, hafner2018learning}). Video frames are first processed with a convolutional encoder module to produce image embeddings $\iota_t = \text{CNN}_{enc}(I_t)$. Inferred frame embeddings $\hat\iota$ are decoded with a convolutional decoder $\hat{I}^n_j = \text{CNN}_{dec}(\hat{\iota}^{n}_j)$. 
The keyframe predictor $p(K^{1:N}, \tau^{1:N} | z^{1:N},  I_{co})$ is parametrized with a Long Short-Term Memory network (LSTM, \cite{hochreiter1997long}). To condition the keyframe predictor %
on past frames, we initialize its state with %
the final state of another LSTM that processes the conditioning frames. %
Similarly, we parametrize the sequence inpainter $p(I_{\tau^n:\tau^{n+1}} |  K^{n}, K^{n+1}, \tau^{n+1}-\tau^n)$ with an LSTM. We condition the inpainting on both keyframe embeddings, $\hat{\kappa}^{n-1}$ and $\hat{\kappa}^{n}$, as well as the temporal offset between the two, $\delta^n$, %
by passing these inputs through a multi-layer perceptron that produces the initial state of the inpainting LSTM. 

We use a Gaussian distribution with diagonal covariance matrix and identity variance as the output distribution for both the keyframe predictor and the inpainting model and a multinomial distribution for $\delta^n$. We parametrize the inference $q(z^{1:N} | I, I_{co})$ with an LSTM with attention over the entire input sequence. The inference distribution is a Gaussian with diagonal covariance matrix, and the prior $p(z^{1:N})$ is a unit Gaussian. 

\begin{figure}
  \centering
  \includegraphics[width=0.7\linewidth]{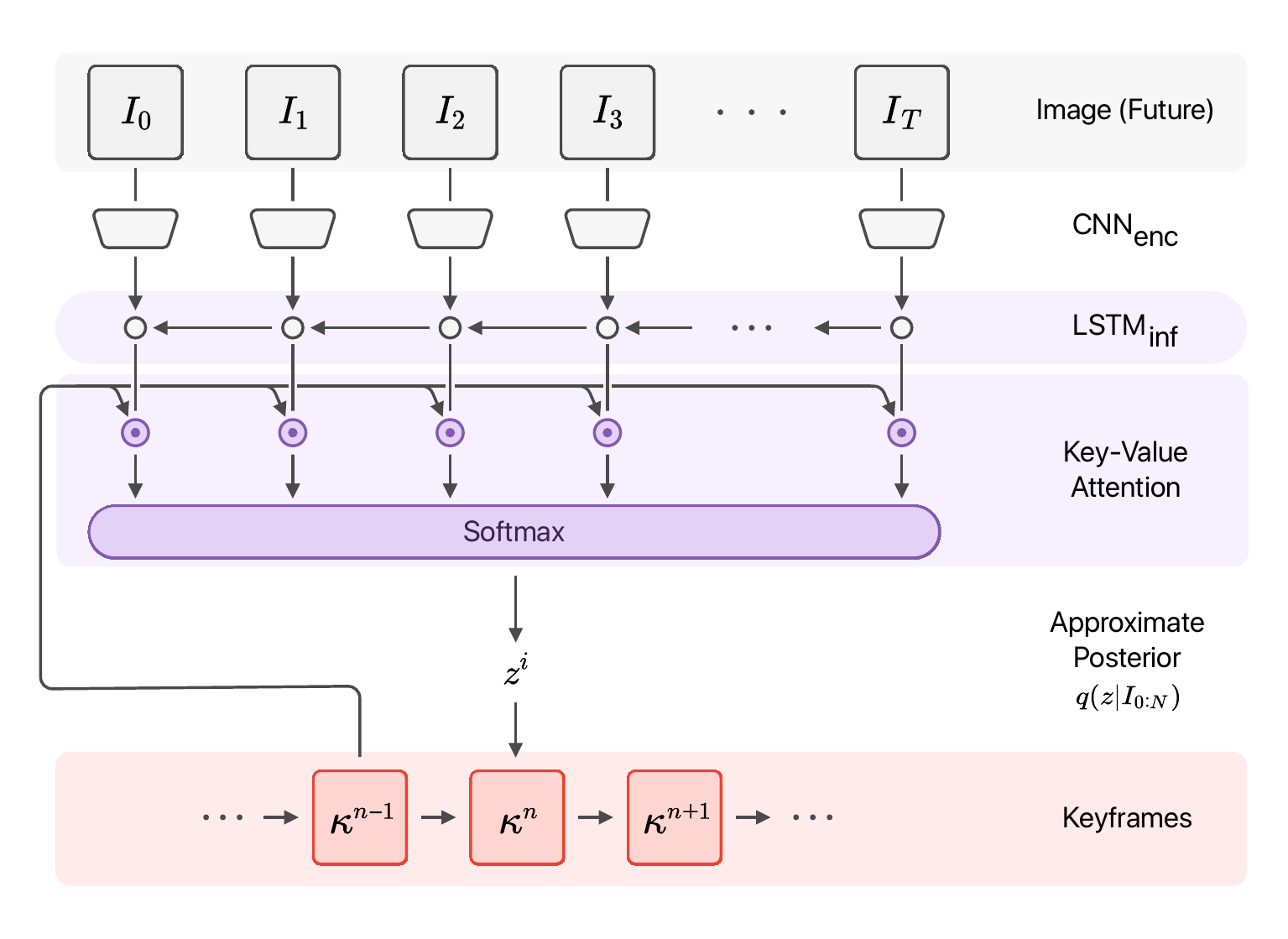}
  \vspace{-20pt}
  \caption{Structure of the keyframe inference network. Depicted is the procedure for inferring the embedding of the $n$-th keyframe, $\kappa^n$, given the previously inferred keyframe embedding $\kappa^{n-1}$ and the future images. The initial state of $\text{LSTM}_{\text{key}}$ is produced by $\text{LSTM}_{\text{cond}}$ (not shown), which takes the embedding of past images as input.
  }
  \label{fig:inference_network}
  \vspace{-15pt}
\end{figure}

We found simple attention over LSTM outputs to be an effective inference procedure. Our approximate inference network $\text{LSTM}_{inf}$ outputs $({\kappa}^{inf}_t, \zeta_t)_{t \leq T}$, where ${\kappa}^{inf}$ is an embedding used to compute an attention weight and the $\zeta_t$ are values to be attended over. We compute the posterior distribution over $z^t$ using a key-value attention mechanism \citep{bahdanau2014neural, luong2015effective}: 
\begin{align}
    a_{n,t} & = \exp(d(\hat{\kappa}^{n-1}, {\kappa}_t^{inf})) \\
    \mu^n, \sigma^n & = (\sum_t a_{n,t} \zeta_t) / \sum_t a_{n,t}.
\end{align} The distance metric, $d$, is the standard inner product. The architecture used for keyframe inference, including the attention mechanism, is depicted in Fig.~\ref{fig:inference_network}.

\subsection{Training Procedure} %

\label{sec:supp_training}

We train our model in two stages. First, we %
train the sequence inpainter to inpaint between ground truth frames sampled with random offsets, thus learning interpolation strategies for a variety of different inputs. In the second stage, we train the keyframe predictor using the loss from Eq.~\ref{eq:loss} by feeding the predicted keyframe embeddings to the inpainter. In this stage, the weights of %
the inpainter are frozen and are only used to backpropagate errors to the rest of the model. We found that this simple two-stage procedure improves optimization of the model.  

We use L1 reconstruction losses to train the keyframe predictor. We found that this and adding a reconstruction loss on the predicted embeddings of the keyframes, weighted with a factor $\beta_\kappa$, %
improved the ability of the model to produce informative keyframes. Target embeddings are computed using the same soft relaxation used for the target keyframes. More details of the loss computation are given in Sec.\ \ref{sec_supp_loss} and Algorithm \ref{alg:loss} of the appendix. %

\section{Experimental setup}
\label{sec_supp_experimental}

We set the prediction horizon to $T=30$ frames and predict $N=6$ segments with $J=10$ frames each for the SBM dataset and $6$ frames each for the Pushing dataset. We pre-train the interpolator on segments of two to eight frames for Structured Brownian motion data, and two to six frames for Pushing data. The weight on the KL-divergence term for the interpolator VAE is $\num{1e-3}$. For training the keyframe predictor, we set $\beta_{K} = 0$, $\beta_{\kappa} = 1, \beta = \num{5e-2}$. The hyperparameters were hand-tuned. We activate the generated images with a sigmoid function and use BCE losses on each color channel to avoid saturation. The convolutional encoder and decoder both have three layers for the Structured Brownian motion dataset and four layers for the Pushing dataset. We use a simple two-layer LSTM with a 256-dimensional state in each layer for all recurrent modules. Each LSTM has a linear projection layer before and after it that projects the observations to and from the correct dimensions. We use the Adam optimizer \citep{kingma2014adam} with $\beta_1 = 0.9$ and $\beta_2 = 0.999$, batch size of 30, and a learning rate of $\num{2e-4}$. Each network was trained on a single high-end NVIDIA GPU. We trained the interpolator for 100K iterations, and the keyframe predictor for 200K iterations. The toal training time took about a day. 

In the Pushing environment, we use a held-out test set of 120 of sequences. The Structured Brownian Motion dataset is generated automatically and is potentially infinite. We used 1000 testing samples on the Structured Brownian Motion generated using a different random seed.

\section{Data collection in the MuJoCo environment}
\label{sec_supp_mjc_data}

The data collection for our pushing dataset was performed in an environment simulated in MuJoCo (\cite{todorov2012mujoco}). In the environment, a robot arm initialized at the center of the table pushes an object to a goal position located at the other side of a wall-shaped obstacle. 

The demonstrations followed a rule-based algorithm that first samples subgoals between the initial position of the object and the goal and then runs a deterministic pushing procedure to the subgoals in order. The ground truth keyframes of the demonstrations were defined by frames at which a subgoal reaching routine was completed. 

We subsampled demonstration videos by a factor of two when saving them to the dataset, dropping every other frame in the trajectory and aggregating actions of every two consecutive frames. We filtered all demonstrations that fail to push the object to the goal position within a predefined horizon. 

\section{Details of the loss computation algorithm}
\label{sec_supp_loss}

\begin{wrapfigure}{R}{0.6\textwidth}
  \centering
  \vspace{-15pt}
  \includegraphics[width=1\linewidth]{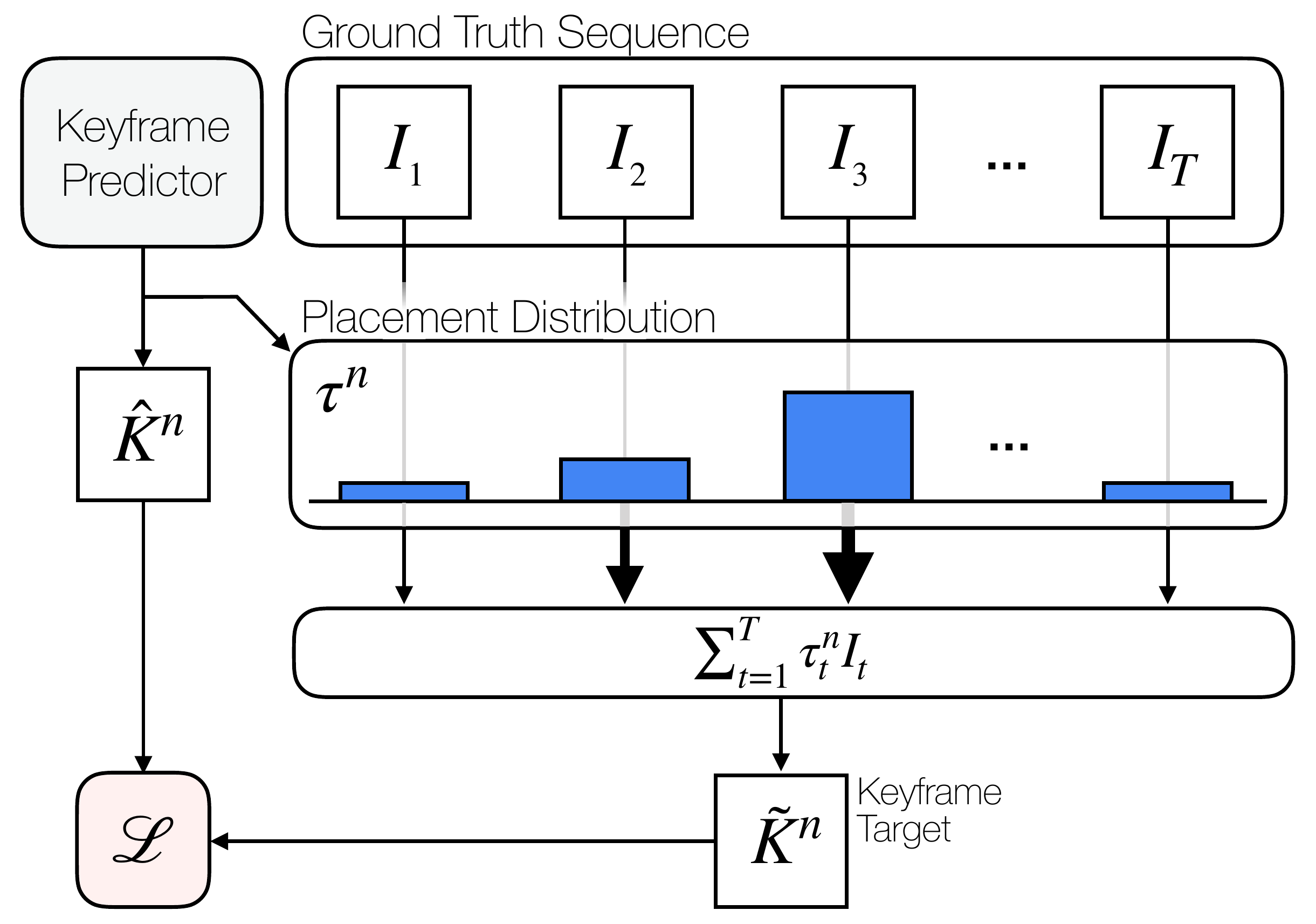}
  \vspace{-15pt}
  \caption{
  Soft keyframe loss in the relaxed formulation. For each predicted keyframe $\hat{K}^{n}$ we compute a target image $\tilde{K}^n$ as the sum of the ground truth images weighted with the corresponding distribution over index $\tau^n$. Finally, we compute the reconstruction loss between the estimated image $\hat{K}^{n}$ and the soft target $\tilde{K}^{n}$. 
}
  \label{fig:loss}
  \vspace{-20pt}
\end{wrapfigure}

We describe the details of the continuous relaxation loss computation in Algorithm \ref{alg:loss}.

Note that we efficiently implement computing of the cumulative distributions $\tau$ as a convolution, which allows us to vectorize much of the computation. The computational complexity of the proposed implementation scales linearly with the number of keyframes $N$, the number of allowed frames per segment $J$ and the number of ground truth frames $T$. The final complexity is $\mathcal{O}(NTJ)$, which we find in practice to be negligible compared to the time needed for the forward and backward pass. 

\begin{algorithm*}[t]
    \caption{Continuous relaxation loss computation}
    \label{alg:loss}
    \begin{algorithmic}
    \STATE {\bfseries Parameters:} $\text{Number of ground truth frames } T, \text{Number of keyframes } N$
    \STATE {\bfseries Input:} $\text{Ground truth frames } I_{1:T}, \text{Generated frames } \hat{I}^t_i, \text{generated offset distributions } \delta^n$
    \STATE Convert distributions of interframe offsets $\delta^n$ to keyframe timesteps $\tau^n$. Set $\tau^1 = {\delta}^1.$
    \FOR{$t = 2\dots M$}
        \STATE %
        $ \tau^n = \tau^{n-1} * \delta^n \text{, i.e. }\tau^n_t = \sum_j \tau^{n-1}_{n-j+1} \delta^n_{j}. $
    \ENDFOR
    \STATE Compute probabilities of keyframes being within the predicted sequence: $c^n = \sum_{t \leq T} \tau^n_{t}$.
    \STATE Compute soft keyframe targets: $\tilde{K}^{n} = \sum_t \tau^n_t I_{t}$.
    \STATE Compute the keyframe loss: $\mathcal{L}_{key} = (\sum_{n} c^n || \hat{K}^n -  \tilde{K}^{n} ||^2) / \sum_{n} c^n.$
    \STATE
    \STATE Get probabilities of segments ending after particular frames: $e^n_j = \sum_{j > i} \delta^n_j$.
    \STATE Get distributions of individual frames' timesteps: $ m^n_{j,t} \propto \tau^{n-1}_{t-j+1} e^n_j$.
    \STATE Compute soft individual frames: $\tilde{I}_t = \sum_{t,i} m^n_{j,t} \hat{I}^{t}_i $
    \STATE Compute total sequence loss: $\mathcal{L}_{total} = \mathcal{L}_{key} + \beta_I \sum_{t} || I_t - \tilde{I}_t ||^2 $.
    
    \end{algorithmic}
\end{algorithm*}

\section{Planning algorithm}
\label{sec_supp_planning}

To apply the \keyin model for planning, we use the approach for visual planning outlined in Algorithm~\ref{alg:latent_planning}. At the initial timestep, we use the cross-entropy method (CEM, \cite{blossom2006cross}) to select subgoals for the task. To do so, we sample $\tilde{M}$ latent sequences $z_0$ from the prior  \(\mathcal{N}(0,I)\) and use the keyframe model to retrieve $\tilde{M}$ corresponding keyframe sequences, each with $L$ frames. We define the cost of an image trajectory as the distance between the target image and the final image of each keyframe sequence defined under a domain-specific distance function. In the update step of the CEM algorithm, we rank the trajectories based on their cost and fit a diagonal Gaussian distribution to the latents $z'$ that generated the \(\tilde{M}^\prime = r\tilde{M}\) best sequences, where $r$ is the elite ratio. We repeat the procedure above for a total of \(N_{it}\) iterations.

After subgoals are selected, we use a CEM based planner to produce rollout trajectories. Similar to the subgoal generation procedure, at each time step, we initially sample \(M\) action sequences $u_0$ from the prior \(\mathcal{N}(0,I)\) and use the ground truth dynamics of the simulator to retrieve \(M\) corresponding image sequences, each with \(l\) frames\footnote{In practice, we clip the sampled actions to a maximal action range $[-a_{\text{max}}, +a_{\text{max}}]$ before passing them to the simulator.}. We define the cost of an image trajectory as the distance between the target image and the final image of each trajectory under the domain-specific distance metric. In the update step, we rank the trajectories based on their cost and fit a diagonal Gaussian distribution to the actions \(u^\prime\) that generated the \(M^\prime = rM\) best sequences. After sampling a new set of actions $u_{n+1}$ from the fitted Gaussian distributions we repeat the procedure above for a total of \(N_{it}\) iterations.

Finally, we execute the first action in the action sequence corresponding to the best rollout of the final CEM iteration. The action at the next time step is chosen using the same procedure with the next observation as input and reinitialized action distributions. The algorithm terminates when the specified maximal number of planning steps \(T_{\text{max}}\) has been executed or the distance to the goal is below a set threshold.

We switch between planned subgoals if (i) the subgoal is reached, i.e. the distance to the subgoal is below a threshold $d_{\text{switch}}$ measured in pixels, or (ii) the current subgoal was not reached for $T_{s, \text{max}}$ execution steps. We use the true goal image as an additional, final subgoal.

\begin{algorithm*}[h]
	\caption{Keyframe-based planning.} 
	\label{alg:latent_planning}
	\begin{algorithmic}
		\STATE {\bfseries Input:} $\text{Keyframe model } \hat{K}_{1:L} = \text{LSTM}_{key}(I, z_{1:L})$, $\text{true dynamics } I_{t+l} = F(I_{t}, u_{t})$, \\$\quad\quad\quad\text{subgoal index update heuristic } \text{ix}_{t + 1} = f(\text{ix}_t, I_t, K_{1:L})$, $\text{start \& goal images } I_1 \text{ and } I_{\text{goal}}.$
		\STATE $\text{Initialize latents from prior: } z_0 \sim \mathcal{N}(0,I).$
		\FOR{$i = 1\dots N_{it}$}
      	  \STATE $\text{Rollout keyframe model for }L\text{ steps, obtain }\tilde{M}\text{ future keyframe sequences } \hat{K}_{1:L}.$
          \STATE $\text{Compute distance between final and goal image: } c = \text{dist}(\hat{K}_{L}, I_{\text{goal}}).$
          \STATE $\text{Choose }\tilde{M}^\prime\text{ best sequences, refit Gaussian distribution: } \mu_{i+1}^{key}, \sigma_{i+1}^{key} = \text{fit}(K_i^{\prime}).$
          \STATE $\text{Sample new latents from updated distribution: }z_{i+1} \sim \mathcal{N}(\mu_{i+1}^{key}, \sigma_{i+1}^{key}).$
        \ENDFOR
        \STATE $\text{Use best sequence of latents to obtain subgoals: } K_{1:L}^{\ast} = \text{LSTM}_{key}(I_1, z_{N_{it}, 0}^{*}).$
        \STATE $\text{Set current subgoal to } \text{ix}_1 = 1.$
        \FOR{$t = 1\dots T_{plan}$}
          \STATE $\text{Perform subgoal update } \text{ix}_{t} = f(\text{ix}_{t - 1}, I_{t - 1}, K_{1:L}^{\ast}).$
          \STATE $\text{Initialize actions from prior: } u_0 \sim \mathcal{N}(0,I).$
          \FOR{$i = 0\dots N_{it}$}
          	\STATE $\text{Rollout true dynamics for }l\text{ steps, obtain }M\text{ future sequences }  I_{t:t+l}.$
            \STATE $\text{Compute distance between final and subgoal image: } c = \text{dist}(I_{t+l}, K_{\text{ix}_{t}}).$
            \STATE $\text{Choose }M^\prime\text{ best sequences, refit Gaussian distribution: } \mu_{i+1},\sigma_{i+1} = \text{fit}(u_i^{\prime}).$
            \STATE $\text{Sample new actions from updated distribution: }u_{i+1} \sim \mathcal{N}(\mu_{i+1}, \sigma_{i+1}).$
          \ENDFOR
          \STATE $\text{Execute }u_{N_{it},0}^{\ast}\text{ and observe next image } I_{t}.$
		\ENDFOR
	\end{algorithmic}
\label{alg:two_stage_planning}
\end{algorithm*}

The parameters used for our visual planning experiments are listed in Table \ref{tab:vs_param}.

\begin{table}
\caption{Hyperparameters for the visual planning experiments.}
  \centering
  \begin{tabular}{ll}
    \toprule
    \multicolumn{2}{c}{planning Parameters}\\
    \midrule
    Max. planning timesteps ($T_{\text{max}}$) & 60  \\
    Max. per subgoal timesteps ($T_{s, \text{max}}$) & 10  \\
    Keyframe prediction horizon ($L$) & 6 \\
    \# keyframe sequences ($\tilde{M}$) & 200  \\
    planning horizon ($l$) & 8 \\
    \# planning sequences ($M$) & 200  \\
    Elite fraction ($r = M^\prime / M$) & 0.05 \\
    \# refit iterations ($N$) & 3 \\
    max. action ($a_{\text{max}}$) & 1.0 \\
    $d_{\text{switch}}$ & 5 \\
    \bottomrule
  \end{tabular} 
\label{tab:vs_param}
\end{table}

\section{Surprise detection}
\label{sec_surprise}

Standard stochastic video prediction methods do not attempt to estimate keyframes, as they are designed to densely estimate future sequences frame-by-frame. Accordingly, they cannot be used directly as baselines for keyframe prediction methods, such as \textsc{KeyIn}. However, \citet{denton2018stochastic} observe that the variance of the learned prior of a stochastic video prediction model tends to spike before an uncertain event happens. We exploit this observation to find the points of high uncertainty for our strong Surprise baseline. We use the KL divergence between the prior and the approximate posterior $\text{KL}[q(z_t| I, I_{co}) || p(z_t)]$ to measure the surprise. This quantity can be interpreted as the number of bits needed to encode the latent variable describing the next state, it will be larger if the next state is more stochastic.

We train a stochastic video prediction network with a fixed prior (SVG-FP, \cite{denton2018stochastic}) with the same architectures of encoder, decoder, and LSTM as our model. We found that selecting the peaks of suprise works the best for finding true keyframes. The procedure we use to select the keyframes is described in Algorithm \ref{alg:surprise}. In order to find the keyframes in a sequence sampled from the prior, we run the inference network on the generated sequence. We provide comparisons to alternative formulations of surprise in Table~\ref{tab:surprise_comp}.

\begin{algorithm*}[t]
    \caption{Selecting keyframes via Surprise}
    \label{alg:surprise}
    \begin{algorithmic}
    \STATE {\bfseries Parameters:} $\text{Number of ground truth frames } N, \text{Desired number of keyframes } M$
    \STATE {\bfseries Input:} $\text{Input sequence } I, \text{Stochastic Video Prediction model } SVG(.)$
    \STATE Run the inference network over the sequence: $q(z_{1:T}|I, I_{co}) = SVG(I, I_{co}).$
    \STATE Get the surprise measure: $s_t = KL[q(z_t| I, I_{co}) || p(z_t)].$
    \STATE Find the set of peak surprise points $S$ where: $s_t > s_{t+1} \land s_t < s_{t-1}.$
    \STATE {\bfseries Return:} The $M$ keyframes from $S$ with maximum surprise.
    \end{algorithmic}
\end{algorithm*}

\begin{table}[t]
\caption{We compare different formulations for a surprise-based keyframe detection method: (1) detecting maxima of the KL divergence between prior and posterior in a stochastic prediction model, (2) detecting maxima in the lower bound on data likelihood $\log p$ (ELBO) of a stochastic prediction model, (3) the formulation proposed in \cite{denton2018stochastic} that detects maxima of the variance of a learned prior distribution.}\label{tab:surprise_comp}
\vspace{-20pt}
\begin{center}
\begin{small}
\begin{sc}
\begin{tabular}{lllllll}
\toprule
\footnotesize
Dataset  &  \multicolumn{3}{c}{Push} &  \multicolumn{3}{c}{Gridworld}\\
\cmidrule(r){2-4} \cmidrule(r){5-7} 
Method & F1 $\uparrow$ & $\min d_{\text{KF}}^{\text{true}} \downarrow$ & $\min d_{\text{KF}}^{\text{pred}} \downarrow$ & F1 $\uparrow$ & $\min d_{\text{KF}}^{\text{true}} \downarrow$ & $\min d_{\text{KF}}^{\text{pred}} \downarrow$\\
\midrule

KL-Suprise
& \bm{$0.25$} & \bm{$1.44$} & \bm{$2.05$} &
$0.32$ & $1.42$	& $1.86$\\

$\log p$-Surprise 
& \bm{$0.24$} & \bm{$1.42$} & \bm{$2.08$} &
$0.31$ & $1.45$ & $1.83$\\

\cite{denton2018stochastic}
& $0.17$ & $1.73$ & $2.01$ & 	
\bm{$0.35$} & \bm{$1.10$} & \bm{$1.53$}\\

\bottomrule
\end{tabular}

\end{sc}
\end{small}
\end{center}
\end{table}

\section{Robustness of Keyframe Detection}
\label{sec:supp_kf_robust}

\begin{table}
\caption{Keyframe discovery for varied number of predicted keyframes. The data has approximately $6$ keyframes. %
Uninterpretable entries are omitted for clarity: see the text for details.}\label{tab:precision_recall}
\begin{center}
\begin{small}
\begin{sc}
\begin{tabular}{l|l|lll}
\toprule
\multicolumn{2}{r|}{\# Keyframes}  & $4$ & $6$ & $8$ \\\hline
\multirow{ 2}{*}{Brownian} & Precision   & $0.92$ & $0.92$ & - \\
 & Recall   & - & $0.96$ & $0.90$ \\\hline
\multirow{ 2}{*}{Push} & Precision   & $0.30$ & $0.38$ & - \\
 & Recall   & - & $0.48$ & $0.46$ \\\hline
 \multirow{ 2}{*}{Gridworld} & Precision   & $0.37$ & $0.43$ & - \\
 & Recall   & - & $0.41$ & $0.40$ \\
\bottomrule
\end{tabular}
\end{sc}
\end{small}
\end{center}
\end{table}

In our experiments we showed that when the sequence can indeed be summarized with $N$ keyframes, \keyin predicts the keyframes that correspond to our notion of salient frames (we show that these results are consistent across metrics in Table~\ref{tab:kf_dist}). However, what happens if we train \keyin to select a larger or a smaller amount of keyframes?

\begin{table}[t]
\caption{In addition to the F1 scores we report the minimal temporal distance to the next keyframe as an additional metric that is more graceful with respect to "close misses". Specifically, we report the distance to the next annotated keyframe averaged across predicted keyframes, $\min d_{\text{KF}}^{\text{true}}$, and, inversely, the distance to the next predicted keyframe for each annotated keyframe, $\min d_{\text{KF}}^{\text{pred}}$. For both datasets the distance metrics support the F1 results: \keyin discovers keyframes that are better aligned with the annotated keyframes than the baselines.}\label{tab:kf_dist}
\vspace{-15pt}
\begin{center}
\begin{small}
\begin{sc}
\begin{tabular}{lllllll}
\toprule
\footnotesize
Dataset  &  \multicolumn{3}{c}{Push} &  \multicolumn{3}{c}{Gridworld}\\
\cmidrule(r){2-4} \cmidrule(r){5-7} 
Method & F1 $\uparrow$ & $\min d_{\text{KF}}^{\text{true}} \downarrow$ & $\min d_{\text{KF}}^{\text{pred}} \downarrow$ & F1 $\uparrow$ & $\min d_{\text{KF}}^{\text{true}} \downarrow$ & $\min d_{\text{KF}}^{\text{pred}} \downarrow$\\
\midrule

Static
& $0.18$ & $1.67$ & \bm{$1.25$} &
$0.25$ & $1.22$ & $1.07$\\

Surprise 
& $0.25$ & $1.44$ & $2.05$ &
$0.32$ & $1.42$	& $1.86$\\

KeyIn (Ours)
& \bm{$0.43$} & \bm{$1.25$} & $1.86$ & 
\bm{$0.42$} & \bm{$1.03$} & \bm{$0.99$}\\

\bottomrule
\end{tabular}

\end{sc}
\end{small}
\end{center}
\end{table}

\begin{figure*}
  \centering
  \vspace{-4pt}
  \includegraphics[width=\textwidth]{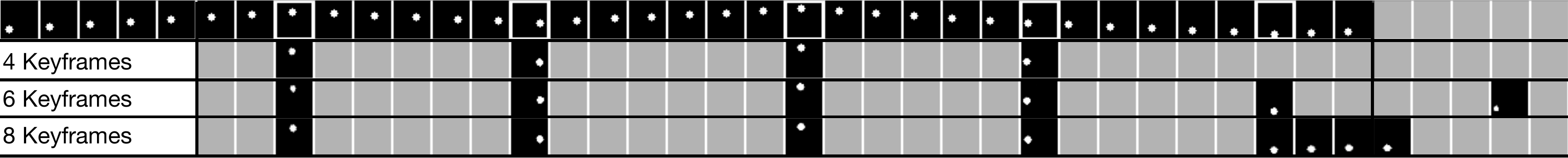}
  \vspace{-20pt}
  \caption{Qualitative keyframe discovery on the Structured Brownian Motion dataset for varying number of predicted keyframes. \textbf{Top}: Ground truth sequence, keyframes with bold white frame. \textbf{Bottom}: \textsc{KeyIn} keyframe predictions at their predicted temporal placement. Even if the number of predicted keyframes does not match the true number of keyframes \textsc{KeyIn} correctly discovers the keyframes and their temporal placement.
  }
  \label{fig:kf_robustness}
   \vspace{-15pt}
\end{figure*}

To evaluate this, we measure \keyin recall with extra and precision with fewer available keyframes. We note that high precision is unachievable in the first case and high recall is unachievable in the second case, since these problems are misspecified. As these numbers are not informative, we do not report them. In Table~\ref{tab:precision_recall}, we see that \keyin is able to find informative keyframes even when $N$ does not exactly match the structure of the data. We further qualitatively show that \keyin selects a superset or a subset of the original keyframes respectively in Fig.\ \ref{fig:kf_robustness}. This underlines that our method's ability to discover keyframe structure is robust to the choice of the number of predicted keyframes.

As a first step towards analyzing the robustness of \keyin under more realistic conditions we report keyframe discovery when trained and tested on sequences with additive Gaussian noise, a noise characteristic commonly found in real-world camera sensors. We find that \keyin is still able to discover the temporal structure on both the Pushing and the Gridworld dataset. For qualitative and quantitative results, see Appendix Fig.~\ref{fig:noise_seqs} and Tab.~\ref{tab:noise_kf_detect}.

\begin{figure*}
  \centering
  \vspace{-4pt}
  \includegraphics[width=\textwidth]{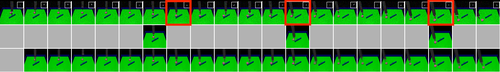}\\
  \vspace{5pt}
  \includegraphics[width=\textwidth]{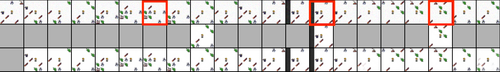}
  \vspace{-20pt}
  \caption{Example keyframe detections on noisy sequences. Red frames mark annotated keyframes. \textbf{Top}: Pushing dataset. \textbf{Bottom}: Gridworld dataset. Each triplet depicts \textit{top}: ground truth sequence with additive Gaussian noise, \textit{middle}: predicted keyframes at the predicted time steps, \textit{bottom}: predicted full sequence. \keyin is reliably able to detect keyframes and reconstruct the full sequence.
  }
  \label{fig:noise_seqs}
   \vspace{-15pt}
\end{figure*}

\begin{table}[t]
\caption{F1 score and distance to closest annotated / predicted keyframe when trained and tested on sequences with additive Gaussian noise. \keyin is able to reliably find keyframes on both datasets even when trained and tested on noisy sequences. Even though the F1 score is lower on the Pushing dataset, the distances indicate that the discovered keyframes are well aligned with the annotated keyframes even under noise.}\label{tab:noise_kf_detect}
\vspace{-20pt}
\begin{center}
\begin{small}
\begin{sc}
\begin{tabular}{lllllll}
\toprule
\footnotesize
Dataset  &  \multicolumn{3}{c}{Push} &  \multicolumn{3}{c}{Gridworld}\\
\cmidrule(r){2-4} \cmidrule(r){5-7} 
Method & F1 $\uparrow$ & $\min d_{\text{KF}}^{\text{true}} \downarrow$ & $\min d_{\text{KF}}^{\text{pred}} \downarrow$ & F1 $\uparrow$ & $\min d_{\text{KF}}^{\text{true}} \downarrow$ & $\min d_{\text{KF}}^{\text{pred}} \downarrow$\\
\midrule
KeyIn, no-noise
& $0.43$ & $1.25$ & $1.86$ & 
$0.42$ & $1.03$ & $0.99$\\

KeyIn, Gauss-noise 
& $0.25$ & $1.21$ & $1.34$ &
$0.43$ & $1.00$ & $0.96$\\

\bottomrule
\end{tabular}

\end{sc}
\end{small}
\end{center}
\end{table}

\section{Video modeling performance}

We further report quantitative results on standard video prediction metrics, Structural Similarity Index (SSIM) and Peak Signal-to-Noise Ratio (PSNR), in Table~\ref{tab:ssim_psnr}. \keyin is able to match performance of two comparable prior approaches, showing that the keyframe-based modeling is able to represent complex data distributions. We also evaluate the quality of \emph{only the keyframes} in Table~\ref{tab:kf_ssim_psnr}, showing that \keyin's adaptively placed keyframes are of higher quality than those predicted by a model with constant keyframe interval.

\begin{table}[t]
\caption{SSIM and PSNR scores on pushing and gridworld dataset. Higher is better.}\label{tab:ssim_psnr}
\begin{center}
\begin{small}
\begin{sc}
\begin{tabular}{lllll}
\toprule
\footnotesize
Dataset  &  \multicolumn{2}{c}{Push} &  \multicolumn{2}{c}{Gridworld}\\
\cmidrule(r){2-3} \cmidrule(r){4-5} 
Method & PSNR & SSIM & PSNR & SSIM\\
\midrule
\citet{denton2018stochastic} 
& $33.3\pm 0.1$ 
& $0.956\pm 0.001$ 
& $29.4\pm 0.1$ 
& $0.812\pm 0.01$\\

Jumpy 
& $33.7\pm 0.1$      
& $0.960\pm 0.001$ 
& $29.9\pm 0.1$ 
& $0.831\pm 0.001$\\

KeyIn (Ours)
& $33.4\pm 0.7$ 
& $0.959\pm 0.001$
& $29.3\pm 0.1$ 
& $0.820\pm 0.001$ \\

\bottomrule
\end{tabular}

\end{sc}
\end{small}
\end{center}
\end{table}

\begin{table}[t]
\caption{\textbf{Keyframe} SSIM and PSNR scores on pushing and gridworld dataset. Higher is better.}\label{tab:kf_ssim_psnr}
\begin{center}
\begin{small}
\begin{sc}
\begin{tabular}{lllll}
\toprule
\footnotesize
Dataset  &  \multicolumn{2}{c}{Push} &  \multicolumn{2}{c}{Gridworld}\\
\cmidrule(r){2-3} \cmidrule(r){4-5} 
Method & PSNR & SSIM & PSNR & SSIM\\
\midrule
Jumpy 
& $28.25\pm 0.11$    	  
& $0.904\pm 0.001$ 
& $16.3\pm 0.17$ 	
& \bm{$0.632\pm 0.001$}\\

KeyIn (Ours)
& \bm{$29.5\pm 0.16$} 	
& \bm{$0.911\pm 0.001$}
& \bm{$18.4\pm 0.17$} 	
& \bm{$0.636\pm 0.001$} \\

\bottomrule
\end{tabular}

\end{sc}
\end{small}
\end{center}
\end{table}

\begin{figure*}
  \centering
  \includegraphics[width=\linewidth]{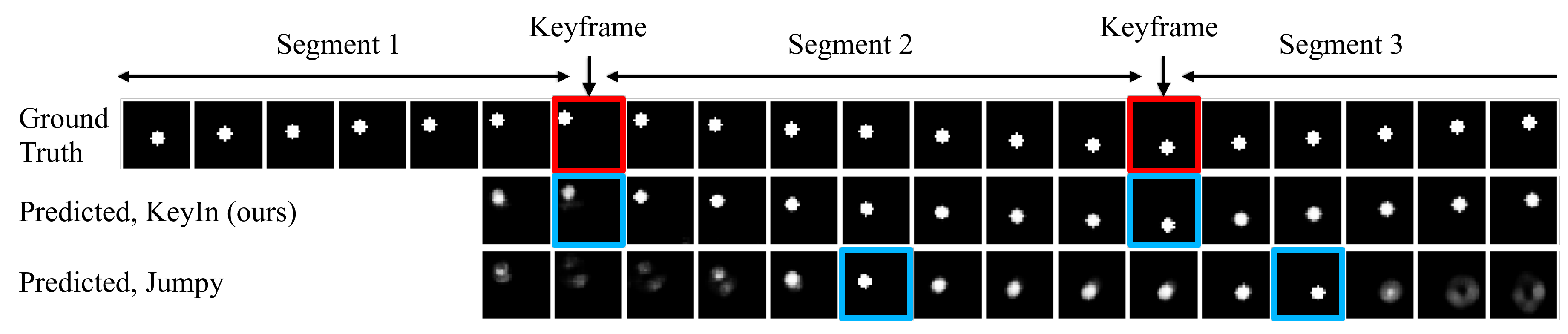}
  \vspace{-15pt}
  \caption{Sequences generated by \textsc{KeyIn} and a method with constant temporal keyframe offset (Jumpy) on Brownian Motion data. Generation is conditioned on the first five frames. The first half of the sequence is shown. Movement direction changes are marked red in the ground truth sequence and predicted keyframes are marked blue. We see that \textsc{KeyIn} can correctly reconstruct the motion as it selects an informative set of keyframes. %
  The sequence generated by the Jumpy method does not reproduce the direction changes since they cannot be inferred from the selected keyframes.
  }
  \label{fig:bb_baseline}
  \vspace{-10pt}
\end{figure*}

\begin{figure}
    \centering
    \begin{minipage}{.5\textwidth}
        \centering
        \includegraphics[width=\linewidth]{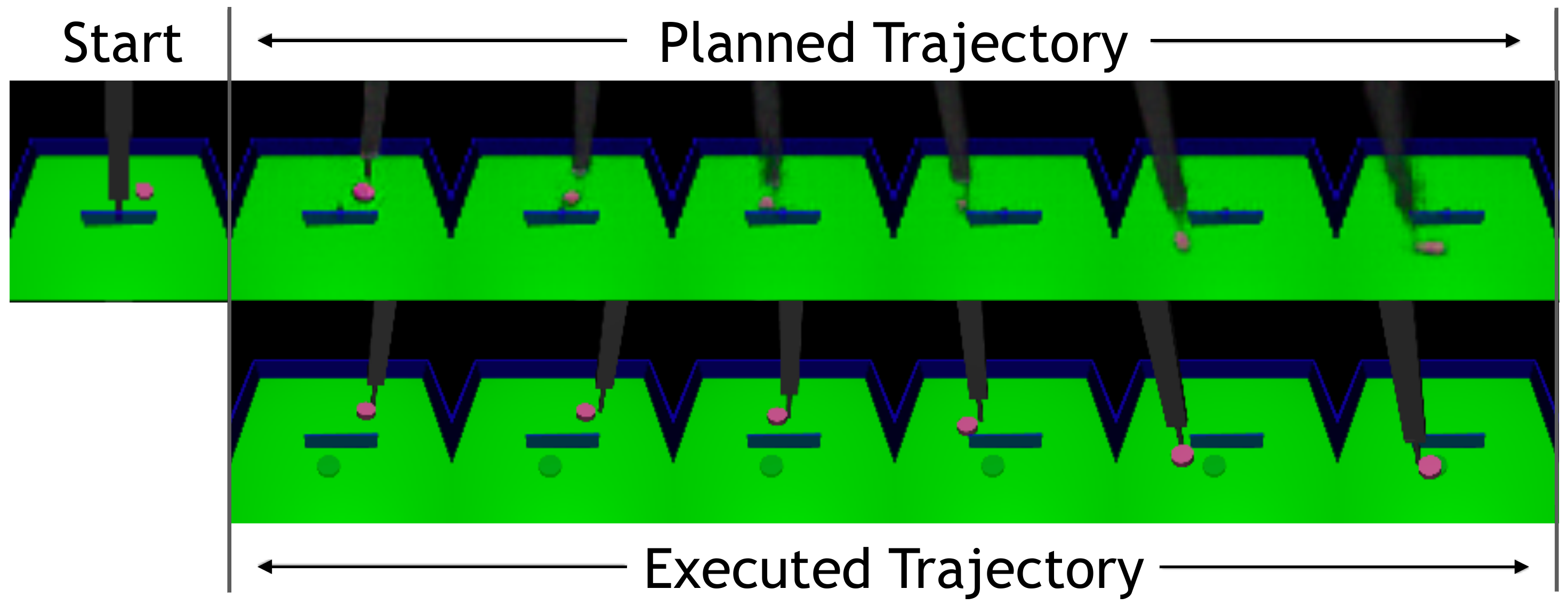}
    \end{minipage}%
    \begin{minipage}{0.5\textwidth}
        \centering
        \includegraphics[width=\linewidth]{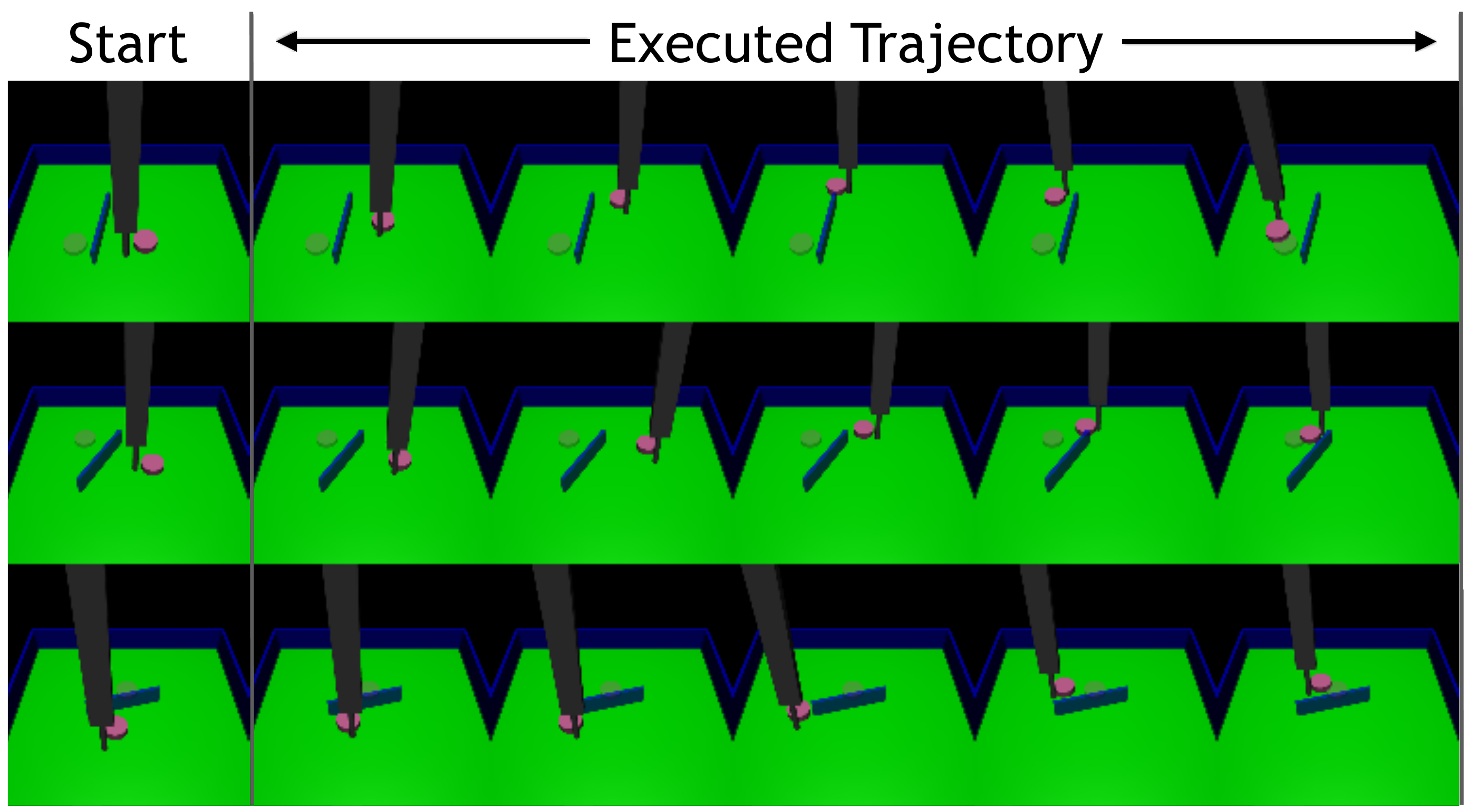}
    \end{minipage}
    \caption{\textbf{Left}: The top row shows the planned subgoals using \textsc{KeyIn}. The bottom row shows snapshots from a successful trajectory between the start state on the left and goal state (depicted transparently in each frame). The low-level controller closely follows the given subgoals and successfully reaches the goal. \textbf{Right}: Additional sample plan executions when planning with \keyin.}
    \label{fig:push_exec_traces}
\end{figure}

\end{document}